\documentclass{article}
\usepackage[final, nonatbib]{nips_2017}  

\usepackage[utf8]{inputenc} 
\usepackage[T1]{fontenc}    
\usepackage{hyperref}       
\usepackage{url}            
\usepackage{booktabs}       
\usepackage{amsfonts}       
\usepackage{amsmath}
\usepackage{amssymb}
\usepackage{nicefrac}       
\usepackage{microtype}      
\usepackage{graphicx}
\usepackage{rotating}
\DeclareMathOperator*{\argmax}{argmax}

\title{Cross-modal Recurrent Models for Weight Objective \\Prediction from Multimodal Time-series Data}
\author{
Petar Veli\v{c}kovi\'{c}$^{1,3}$, Laurynas Karazija$^1$, Nicholas D. Lane$^{2,3}$, Sourav Bhattacharya$^3$,\\ {\bf Edgar Liberis$^1$, Pietro Li\`{o}$^1$, Angela Chieh$^4$, Otmane Bellahsen$^4$, Matthieu Vegreville$^4$} \\
$^1$University of Cambridge \quad $^2$University of Oxford\\ $^3$Nokia Bell Labs \quad $^4$Nokia Digital Health - Withings \\
}

\begin{document}

\maketitle
\begin{abstract}
We analyse multimodal time-series data corresponding to weight, sleep and steps measurements. We focus on predicting whether a user will successfully achieve his/her weight objective. For this, we design several deep long short-term memory (LSTM) architectures, including a novel \emph{cross-modal LSTM} (\textbf{X-LSTM}), and demonstrate their superiority over baseline approaches. The X-LSTM improves parameter efficiency by processing each modality separately and allowing for information flow between them by way of \emph{recurrent cross-connections}. We present a general hyperparameter optimisation technique for X-LSTMs, which allows us to significantly improve on the LSTM and a prior state-of-the-art cross-modal approach, using a comparable number of parameters. Finally, we visualise the model's predictions, revealing implications about latent variables in this task.
\end{abstract}

\section{Introduction}

Recently, consumer-grade health devices, such as wearables and smart home appliances became more widespread, which presents new data modelling opportunities. Here, we investigate one such task---predicting the users' future body weight in relation to their weight goal given historical weight, and sleep and steps measurements. This study is enabled by a first-of-its-kind dataset of fitness measurements from \emph{$\sim$15000 users}. Data are captured from different sources, such as smartwatches, wrist- and hip-mounted wearables, smartphone applications and smart bathroom scales.



In this work, we show that that deep long short-term memory (LSTM) \cite{hochreiter1997long} models \emph{are} able to produce accurate predictions in this setting, significantly outperforming baseline approaches, even though some factors are only observed latently. We also discover interesting patterns in input sequences that push the network's confidence in success or failure to extremes. We hypothesise that these patterns affect latent variables and link our hypotheses to existing research on sleep.

Most importantly, we improve the parameter efficiency of LSTM models for multimodal input (in this case sleep/steps/weight measurements) by proposing \emph{cross-modal LSTMs (X-LSTMs)}. X-LSTMs extract features from each modality separately, while still allowing for \emph{information flow} between the different modalities by way of \emph{cross-connections}.
Our findings are supported by a general data-driven methodology (applicable to \emph{arbitrary} multimodal problems) that exploits unimodal predictive power to vastly simplify finding appropriate hyperparameters for X-LSTMs (reducing most of the tuning effort to a \emph{single parameter}). We also compare our model to a previous state-of-the-art cross-modal sequential data technique \cite{Ren2016}, outlining its limitations and successfully outperforming it on this task.

\section{Dataset and Preprocessing}\label{secdata}

We performed our investigation on anonymised data obtained from bathroom scales and wearables of the \emph{Nokia Digital Health - Withings} range, gathered using the Withings smartphone application.

The data was pre-processed to remove outliers or users with too few, or too sporadic, data observations. We consider a weight objective \emph{achieved} if there exists a weight measurement in the future that reaches or exceeds it, and \emph{failed} if the user \emph{stops recording weights} (allowing for a long enough window after the end of the recorded sequence) or \emph{sets a more conservative objective}. Following best practices, data are normalised to have mean zero and standard deviation one per-feature.

The derived dataset spans 18036 sequences associated with weight objectives. All of the sequences are comprised of user-related features: height, gender, age category, weight objective and whether it was achieved; along with sequential features---for each day: duration of light and deep sleep, time to fall asleep and time spent awake; number of times awoken during the night; time required to wake up; bed-in/bed-out times; steps and (average) weights for the day. We consider sequences that span at least 10 contiguous days. The dataset contains 6313 successful and 11723 unsuccessful examples.

%


\section{Models under consideration}


\subsection{Baseline models}

We compared deep recurrent models against several common baseline approaches to time-series classification, as outlined in \cite{xing2010brief}. We considered: \emph{Support Vector Machines} (SVMs) with the RBF kernel, \emph{Random Forests} (RFs), \emph{Gaussian Hidden Markov Models} (GHMMs) {and (feedforward) \emph{Deep Neural Networks}} (DNNs).
The hyperparameters have been optimised using a thorough sweep.



\subsection{Long short-term memory}\label{s5}

Our models are based on the LSTM \cite{hochreiter1997long} model, defined as follows for a single cell (similar to \cite{graves2013generating}):
\begin{eqnarray}
	i_t &=& \tanh({\bf W}^{\mathrm{xi}}x_t + {\bf W}^{\mathrm{yi}}y_{t-1} + b^{\mathrm{i}})\label{firsteq}\\
	\{j, f, o\}_t &=& \tilde{\sigma}({\bf W}^{\mathrm{x\{j, f, o\}}}x_t + {\bf W}^{\mathrm{y\{j, f, o\}}}y_{t-1} + b^{\mathrm{\{j, f, o\}}})\label{fourtheq}\\
	c_t &=& c_{t-1} \otimes f_t + i_t \otimes j_t\\
	y_t &=& \tanh(c_t) \otimes o_t \label{lasteq}
\end{eqnarray}
Here, $\bf W^*$ and $b^*$ correspond to weights and biases of the LSTM layer, respectively, and $\otimes$ corresponds to element-wise vector multiplication. $\tanh$ is the hyperbolic tangent, and $\tilde{\sigma}$ is the \emph{hard sigmoid function}. For the remainder of the description, we compress Eqn-s \ref{firsteq}--\ref{lasteq} into $\mathrm{LSTM}(\vec{x}) = \vec{y}$.

Our primary architecture is a 3-layer LSTM model (21, 42 and 84 features) for processing the sequential data. The features computed by the final LSTM layer are concatenated with the height, gender, age category and weight objective, providing the following feature representation:
\begin{equation}
	\mathrm{LSTM}(\mathrm{LSTM}(\mathrm{LSTM}(\vec{\mathrm{wt}} || \vec{\mathrm{sl}} || \vec{\mathrm{st}})))_T || \mathrm{ht} || \mathrm{gdr} || \mathrm{age} || \mathrm{obj}
\end{equation}
where $\vec{\mathrm{wt}}$, $\vec{\mathrm{sl}}$ and $\vec{\mathrm{st}}$ are the input features (for weight, sleep and steps, respectively), $||$ is featurewise concatenation, and $T$ is the length of the initial sequence.
The result is processed by a 3-layer fully-connected network (128, 64, 1 neurons) with logistic sigmoid activation at the very end.


\subsection{Cross-modal LSTM (X-LSTM)}

For this task we also propose a \emph{novel cross-modal} LSTM (\emph{X-LSTM}) architecture which exploits the \emph{multimodality} of the input data explicitly, while using the same number of parameters as the traditional LSTM. We partition the input sequence into three parts (sleep, weight and steps data), and pass \emph{each of those} through a separate three-layer LSTM stream. We also allow for \emph{information flow} between the streams in the second layer, by way of \emph{cross-connections}, where features from a single sequence stream are passed and concatenated with features from another sequence stream (after being passed through an additional LSTM layer). In equation form, outputs of the three streams are:
\begin{eqnarray}\label{e1}
	\vec{h}^\mathrm{\{wt,sl,st\}}_1 &=& \mathrm{LSTM}(\{\vec{\mathrm{wt}}, \vec{\mathrm{sl}}, \vec{\mathrm{st}}\})\\
	\vec{h}^\mathrm{\{wt\rightarrow wt, sl\rightarrow sl, st\rightarrow st\}}_2 &=& \mathrm{LSTM}(\{\vec{h}^\mathrm{wt}_1, \vec{h}^\mathrm{sl}_1, \vec{h}^\mathrm{st}_1\})\\
	\vec{h}^\mathrm{\{wt\rightsquigarrow sl, wt\rightsquigarrow st\}}_2 &=& \mathrm{LSTM}(\{\vec{h}^\mathrm{wt}_1, \vec{h}^\mathrm{wt}_1\})\\
	\vec{h}^\mathrm{\{sl\rightsquigarrow wt, sl\rightsquigarrow st\}}_2 &=& \mathrm{LSTM}(\{\vec{h}^\mathrm{sl}_1, \vec{h}^\mathrm{sl}_1\})\\
	\vec{h}^\mathrm{\{st\rightsquigarrow wt, st\rightsquigarrow sl\}}_2 &=& \mathrm{LSTM}(\{\vec{h}^\mathrm{st}_1, \vec{h}^\mathrm{st}_1\})
\end{eqnarray}
\begin{eqnarray}
	\vec{h}^\mathrm{wt}_3 &=& \mathrm{LSTM}(\vec{h}^\mathrm{wt\rightarrow wt}_2 || \vec{h}^\mathrm{sl\rightsquigarrow wt}_2 || \vec{h}^\mathrm{st\rightsquigarrow wt}_2)\\
	\vec{h}^\mathrm{sl}_3 &=& \mathrm{LSTM}(\vec{h}^\mathrm{sl\rightarrow sl}_2 || \vec{h}^\mathrm{wt\rightsquigarrow sl}_2 || \vec{h}^\mathrm{st\rightsquigarrow sl}_2)\\
	\vec{h}^\mathrm{st}_3 &=& \mathrm{LSTM}(\vec{h}^\mathrm{st\rightarrow st}_2 || \vec{h}^\mathrm{wt\rightsquigarrow st}_2 || \vec{h}^\mathrm{sl\rightsquigarrow st}_2)	\label{e2}
\end{eqnarray}

We used $\vec{h}^{\{x, y, z\}}_2 = \mathrm{LSTM}(\{a, b, c\})$ to denote $\vec{h}^x_2 = \mathrm{LSTM}(a), \vec{h}^y_2 = \mathrm{LSTM}(b), \vec{h}^z_2 = \mathrm{LSTM}(c)$.

Finally, the final LSTM frames across all of the three streams are concatenated before being passed on to the fully-connected classifier: $(\vec{h}^\mathrm{wt}_3 || \vec{h}^\mathrm{sl}_3 || \vec{h}^\mathrm{st}_3)_T || \mathrm{ht} || \mathrm{gdr} || \mathrm{age} || \mathrm{obj}$.

\begin{figure}[tbh]
	\centering
  \includegraphics[height=0.6\linewidth,angle=90,origin=c]{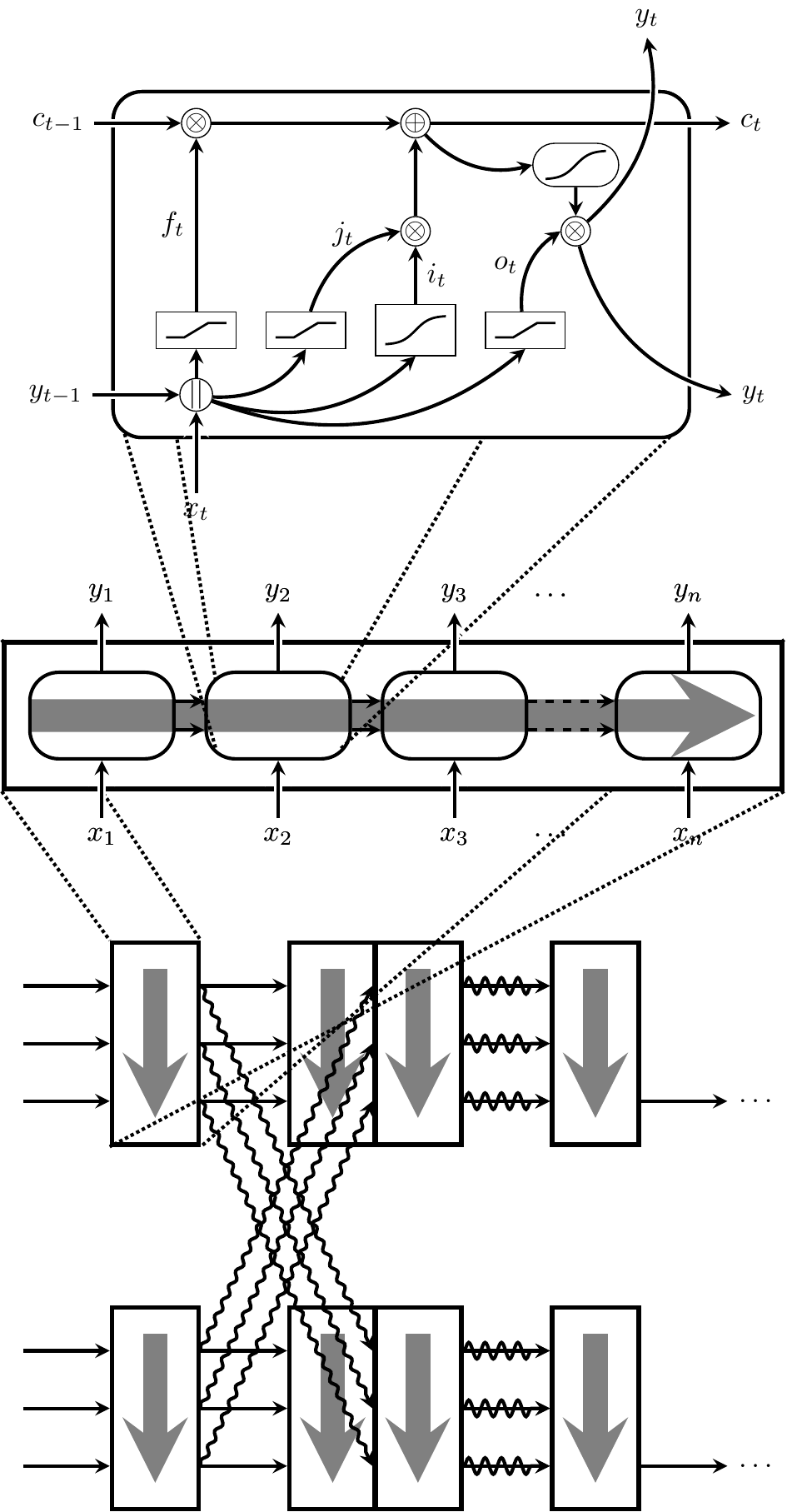}
	\vspace{-20mm} 
	\caption{A diagram of a 3-layer X-LSTM model and one cross-connection in the second layer. \textbf{Left:} A single LSTM block, \textbf{Middle:} An LSTM layer (replicated cell). \textbf{Right:} A 3-layer cross-modal LSTM model with 2 streams. In the second layer, the hidden sequences are passed through a separate LSTM layer and feature-wise concatenated with the main stream sequence to facilitate sharing.}
  \label{figxlstm}
\end{figure}

The illustration of the entire construction process from individual building blocks is {shown in} Fig. \ref{figxlstm}.
Similar techniques have already been successfully applied for handling sparsity within convolutional neural networks \cite{velivckovic2016x} and audiovisual data integration \cite{cangea2017xflow}.
We evaluate \emph{three} cross-connecting strategies: one given by Eqn-s \ref{e1}--\ref{e2} (\emph{A}), one where cross-connections do not have intra-layer LSTMs (\emph{B}), and one without cross-connections (\emph{N}). The latter corresponds to prior work on multimodal deep learning \cite{ngiam2011multimodal,srivastava2012multimodal} and allows for computing the largest number of features within the parameter budget out of all three variants---no parameters are spent on cross-connections.

Finally, we consider a recent state-of-the-art approach for processing multimodal sequential data \cite{Ren2016} which imposes cross-modality via weight sharing ($\mathbf{W}^{y*}$ in Eqn-s \ref{firsteq}--\ref{fourtheq})---we refer to this method as SH-LSTM. This hinders expressivity---in order to share the weights, the matrices to have be of the same size, requiring all modality streams to compute the \emph{same number of features} at each depth level. Keeping the parameter count comparable to the baseline LSTM, we evaluate three strategies for weight sharing: sharing across all modalities (ALL) and sharing only across weight \& sleep, with (WSL) and without (CUT) steps data. This has been informed by the fact that the weight and sleep data have, on their own, been found to be significantly more influential than steps data.

\vspace{1cm}


\subsection{X-LSTM hyperparameter tuning}

In practice, we anticipate X-LSTMs to be derived from a baseline LSTM, in order to redistribute its parameters more efficiently. However, X-LSTMs might introduce an overwhelming amount of hyperparameters to tune. To make the process less taxing, we focus on the \emph{meaning} of the feature counts---their comparative values are supposed to track the \emph{relative significance} of each modality. First, we attempt to solve the task with a basic LSTM architecture using \emph{only one of the modalities}. When scores (e.g. accuracies or AUC) $s_{wt}, s_{sl}$ and $s_{st}$ are obtained for all three modalities, we redistribute the intra-layer feature counts of the X-LSTM according to the ratio $s_{wt} : s_{sl} : s_{st}$. 

To enforce larger discrepancies, we raise the obtained scores to a power $k$. This controls the tendency of the network to favour the most predictive modality when redistributing features. For a fixed choice of $k$, we solve a system of equations in order to derive feature counts for all the intra-layer LSTM layers in an X-LSTM. Thus, most of the effort amounts to finding just \emph{one} hyperparameter---$k$. 

\section{Results}

\subsection{Weight objective success classification}

We performed stratified 10-fold crossvalidation on the baseline classifiers and the proposed LSTM models.
We use \textbf{ROC curves} (and the \textbf{AUC}) as our evaluation metric, but we also report the accuracy, precision, recall, F$_1$ score and the MCC \cite{matthews1975comparison} for the threshold which maximises the F$_1$ score.

To construct competitive X-LSTMs, we computed the AUCs of the individual unimodal LSTMs on a validation dataset. The results were too similar to reliably generate non-uniform X-LSTMs, so we searched for parameter $k$. The X-LSTM performed the best with $k = 30$, and (B) cross-connections (75089 parameters)---we compare it directly with the LSTM (76377 parameters) and the SH-LSTMs.

To confirm that the advantages of our methodology are statistically significant, we have performed paired $t$-testing on the metrics of individual cross-validation folds, choosing a significance threshold of $p < 0.05$.
The SH-LSTM performed the best in its (WSL) variant but even then was unable to outperform the baseline LSTM---highlighting how essential is the ability to accurately specify relative importances between modalities. The results are summarised in Table \ref{tblbaseline}.

\begin{table*}[htb]
\centering
\small{
\begin{tabular}{l c c c c c c c} \toprule
{\bf Metric} & SVM & RF & GHMM & DNN & LSTM & SH-LSTM & X-LSTM\\ \midrule
Accuracy & 67.65\% &	70.97\% &	66.31\% &	68.93\% & 79.12\% & 78.49\% & {\bf 80.30\%} \\
Precision & 52.54\% &	56.05\% &	51.26\% &	53.80\%	& 67.25\% & 65.31\% & {\bf 68.66\%} \\
Recall & 81.02\% &	81.34\% &	82.32\% &	{\bf 83.02\%} &	79.30\% & 82.95\% & 81.62\% \\
F$_1$ score & 63.71\% &	66.25\% &	63.11\% &	65.18\% &	72.69\% & 72.98\% & {\bf 74.37\%} \\
MCC & 39.74\% &	44.75\% &	38.57\% &	42.63\% &	 56.60\% & 56.80\% & {\bf 59.45\%} \\ \midrule
\underline{ROC AUC}& 76.77\% & 79.97\% & 74.86\% &	78.54\%	& 86.91\% & 86.63\% & {\bf 88.07\%}\\
$p$-value & \underline{$2 \cdot 10^{-12}$} & \underline{$6 \cdot 10^{-10}$} & \underline{$7 \cdot 10^{-11}$} & \underline{$2 \cdot 10^{-11}$} & \underline{$1 \cdot 10^{-4}$} & \underline{$4 \cdot 10^{-5}$} & ---\\\bottomrule
\end{tabular}
}
\caption{Comparative evaluation results of the baseline models against the LSTMs after 10-fold crossvalidation. Reported X-LSTM is (B) with $k=30$ and SH-LSTM is (WSL). Reported $p$-values are for the X-LSTM vs. each baseline for the ROC-AUC metric.}
\label{tblbaseline}
\end{table*}

\subsection{Visualising detected features}

\begin{figure}[htb]
\centering
\begin{minipage}{0.45\textwidth}
	\centering
	\includegraphics[width=0.7\textwidth]{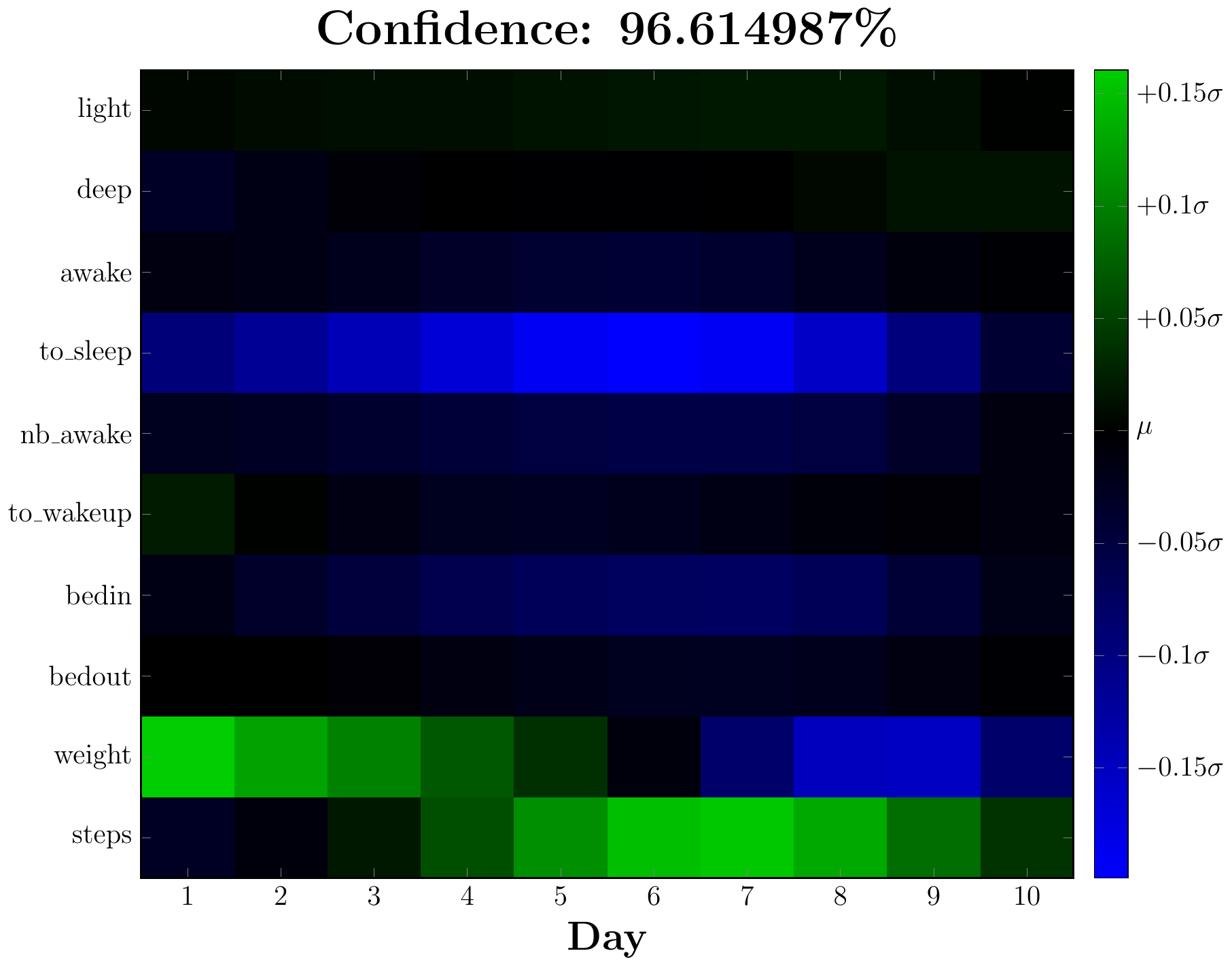}
\end{minipage}
\quad
\begin{minipage}{0.45\textwidth}
	\centering
	\includegraphics[width=0.7\textwidth]{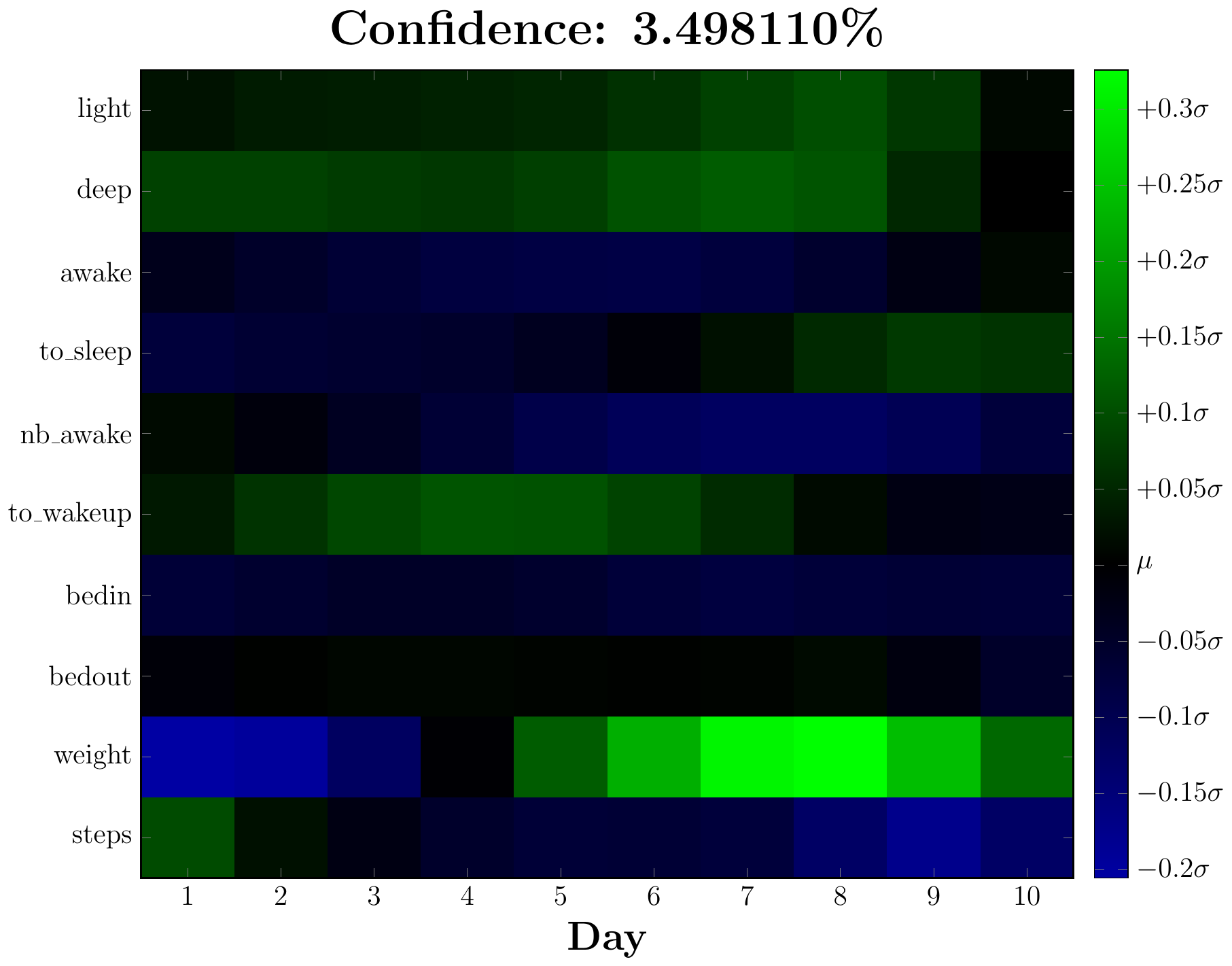}
\end{minipage}
\caption{Iteratively produced artificial sequences that maximise the model's confidence in achieving (\textbf{left}) or failing (\textbf{right}) a $-4\mathrm{kg}$ weight objective. Best viewed in colour.}
\label{figdream}
\end{figure}


It is hard to interpet the parameters of a network directly, so instead we focus on \emph{generating artificial sequences that maximise the network's confidence in success or failure} \cite{simonyan2013deep}.
%
%
Iteratively, we produce an input $\mathbf{I'}$ that maximises the network's confidence, starting from $\mathbf{I}_0 = \mathbf{0}$:
$\mathbf{I'} = \argmax_{\mathbf{I}} \Sigma(\mathbf{I}) - \lambda||\mathbf{I}||^2_2$
where $\Sigma(\mathbf{I})$ is the network's output for $\mathbf{I}$, and $\lambda$ is an $L_2$-regularisation parameter (to penalise large day-to-day variances). We found that $\lambda = 5$ works best. 

Generated sequences spanning 10 days are shown in Fig. \ref{figdream}. As expected, we observe that a user is likely to hit their weight objective if there is a downwards the trend in weight and an upwards trend in steps, and vice-versa for a failing sequence.
Interestingly, the model also uncovered that to have a higher confidence of success, it is important for the user to \emph{fall asleep quicker once going to bed}. This is likely encoding important \emph{latent variables} that we can not directly access from the dataset---for example, a person that takes more time to fall asleep is more likely to snack in the evening, which is known to be detrimental to weight loss (as previously observed in biomedical research~\cite{nedeltcheva2009sleep,sato2011midpoint,kleiser2017sleep}).

\bibliographystyle{plain}
\bibliography{acmlarge-sample-bibfile}

\appendix

\section{Appendices to sections}

In the following sections, we augment the exposition of the main body of our paper to include further relevant details---for the purposes of gaining a better understanding of the utilised dataset, the implemented models, and the presented results.

\subsection{Dataset and preprocessing}

We performed our investigation on anonymised data obtained from several devices across the \emph{Nokia Digital Health - Withings} range. The dataset contains weight, height, sleep and steps measurements, as well as user specified weight objectives. 
Weights are measured by the Withings scale. 
All other data {are} obtained from the Withings application through {the} use of wearables.


Users were first included in the dataset under the condition of having recorded at least 10 {weight} measurements over a 2-month period. In total, the dataset contains 1 664 877 such users. Further processing was performed to remove outliers or those users with too few, or too sporadic, data observations; after this stage $\sim$ 15K users were remaining. The precise steps taken to reach this final dataset are enumerated below.


Obvious {outliers}, reporting {unrealistic heights} (below 130cm or above 225cm), and/or consistent weight {changes} of more than 1.5kg per day have been discarded. 
Steps and sleep are recorded on a per-day basis, while weights are recorded at the user's discretion; to align {the} weight measurements with the other two modalities, we {have} applied a moving average to the person's recorded weight throughout an individual day.
A sequence may be labelled with any weight objective that has been set by the user, and is still unachieved, by the time the sequence ends. {Overly} ambitious objectives (over $\pm$20 kilograms proposed) are ignored.
We consider a weight objective \emph{successful} if there exists a weight measurement in the future that reaches or exceeds it, and we consider it unsuccessful if the user \emph{stops recording weights} (allowing for a long enough window after the end of the recorded sequence) or \emph{sets a more conservative objective} in the meantime.
In line with known best practices in deep learning, data {are} normalised to have mean zero and standard deviation one per-feature.

The derived dataset spans 18036 sequences associated with weight objectives. All of the sequences are comprised of user-related features: height, gender, age category, weight objective; along with sequential features---for each day: duration of light and deep sleep, time to fall asleep and time spent awake; number of times awoken during the night; time required to wake up; bed-in/bed-out times; steps and (average) weights for the day. We consider sequences that span at least 10 contiguous days.



{Every sequence also has a boolean \emph{label}, indicating whether the objective has been successfully achieved at some point in the future. Within our dataset, 6313 of the sequences represent successful examples, while the remaining 11723 represent examples of failure. To address the potential issues of class imbalance, appropriate class weights are applied to all optimisation targets and loss functions. 

\begin{figure}
\centering
\includegraphics[width=0.49\linewidth]{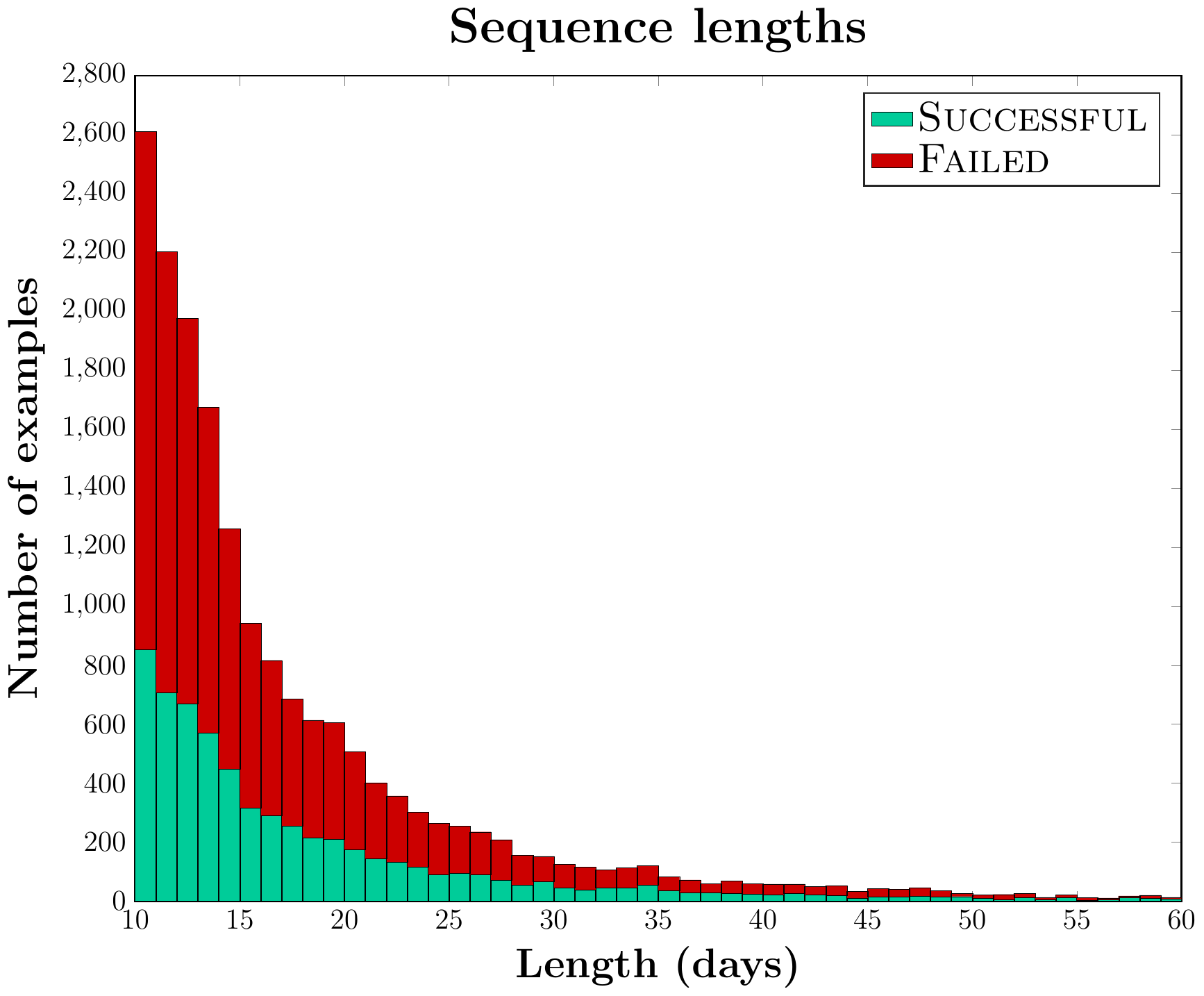}
\includegraphics[width=0.49\linewidth]{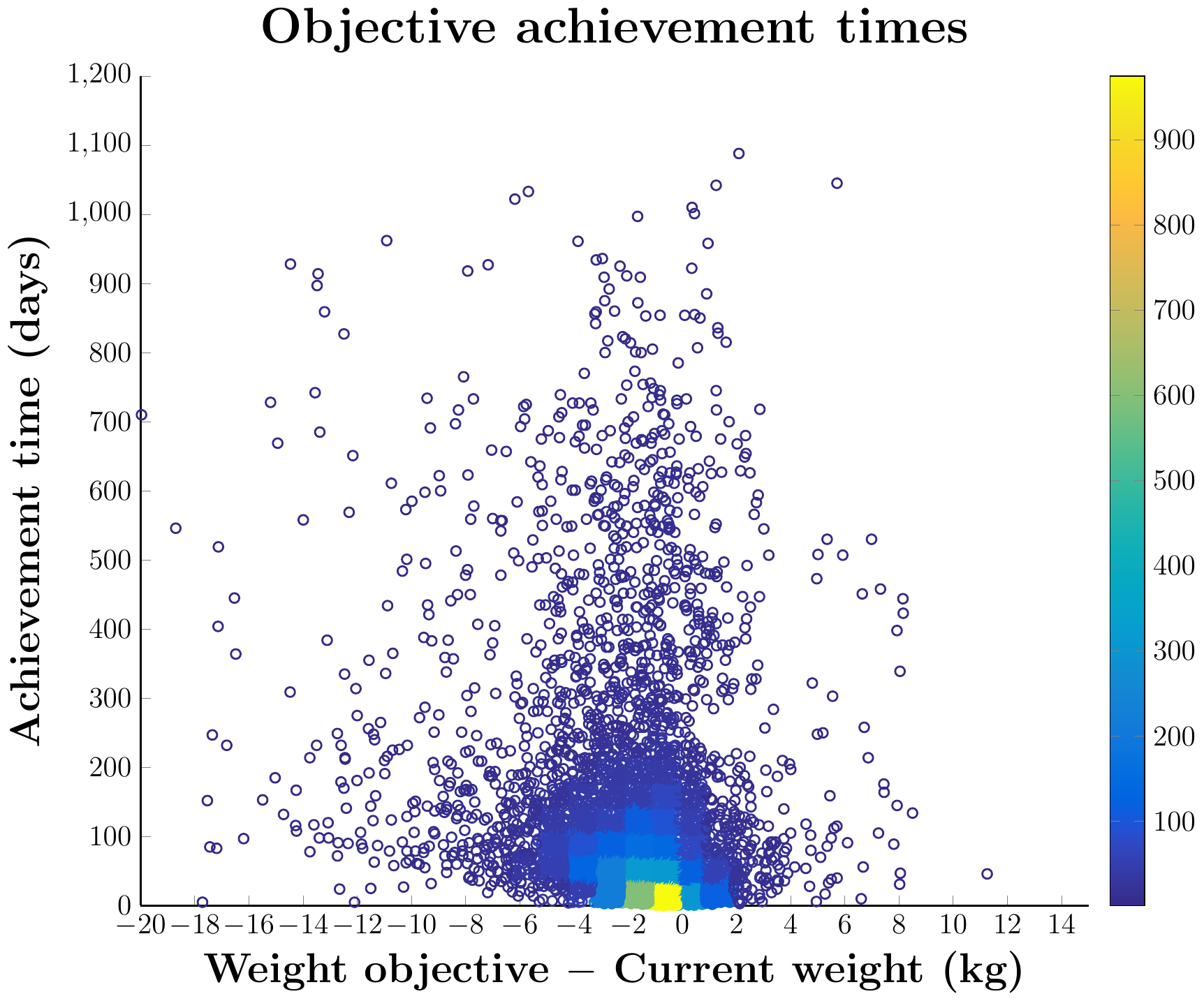}
\caption{{\bf Left:} Plot of the sequence length distribution in the final dataset. {\bf Right:} Mixed heatmap/scatter plot of the weight objectives against their achievement times, for the successful sequences in the final dataset.}
\label{figdataset}
\end{figure}

{In order to get an impression of the statistics present within the dataset, we have generated plots of the sequence length distributions (outliers removed for visibility), as well as scatter plots of successful weight objective magnitudes against their achievement times. These are provided by Figure }\ref{figdataset}.

We perform a task of \emph{probabilistic classification} on the filtered dataset: predicting success for the weight objective, evaluated using crossvalidation (this corresponds to a typical \emph{binary classification problem}).

\subsection{Baseline models}

In order to ascertain the suitability of deep recurrent models on this task, we have compared them on the objective classification task against several common baseline approaches to time-series classification, as outlined in \cite{xing2010brief}. For this purpose, we have considered {four} such models: \emph{Support Vector Machines} (SVMs) using the RBF kernel, \emph{Random Forests} (RFs), \emph{Gaussian Hidden Markov Models} (GHMMs) {and (feedforward) \emph{Deep Neural Networks}} (DNNs).
{The hyperparameters associated with the baseline models have been optimised with a thorough hyperparameter sweep---on a separate validation set---as detailed below.}

For the SVM, we have performed a grid search on its two hyperparameters ($C$ and $\gamma$) in the range $\gamma \in 2^{[-15, 5]}, C \in 2^{[-5, 15]}$, finding the values of $\gamma = 2^{-13}$ and $C = 2^9$ to work best. {For the RF, we have performed a search on the number of trees to use in the range $N \in [10, 100]$, finding $N = 50$ to work best.} {For the GHMM, we have performed a search on the number of nodes to use in the range $N \in [3, 40]$, finding $N = 7$ to work best.} {For the DNN, we have optimised the number of hidden layers (keeping the number of parameters comparable to the recurrent models) in the range $\ell \in [1, 10]$, finding $\ell = 5$ to work best. This implied that each hidden layer had $N = 120$ neurons. All hidden layers apply the \emph{rectified linear} (ReLU) activation} \cite{nair2010rectified}{, and are regularised using batch normalisation} \cite{ioffe2015batch} {and dropout} \cite{srivastava2014dropout} {with $p = 0.5$. All other relevant hyperparameters (such as the SGD optimiser and batch size) are the same as for the recurrent models.}

{For all the non-sequential models (SVM, RF, DNN), we have performed a search on the number of most recent time steps to use in the range $l \in [5, 100]$, finding $l = 10$ to perform the best. The SVM model has been augmented to produce probabilistic predictions (and thus enable its ROC-AUC metric to be computed) by leveraging Platt scaling \cite{Platt99probabilisticoutputs}. }

\subsection{Long short-term memory}\label{s5}

All of our models are based on the long short-term memory (LSTM) \cite{hochreiter1997long} recurrent model. The equations describing a single LSTM cell that we employed (similar to \cite{graves2013generating}) are as follows:
\begin{eqnarray}
	i_t &=& \tanh({\bf W}^{\mathrm{xi}}x_t + {\bf W}^{\mathrm{yi}}y_{t-1} + b^{\mathrm{i}})\label{firsteq}\\
	j_t &=& \tilde{\sigma}({\bf W}^{\mathrm{xj}}x_t + {\bf W}^{\mathrm{yj}}y_{t-1} + b^{\mathrm{j}})\\
	f_t &=& \tilde{\sigma}({\bf W}^{\mathrm{xf}}x_t + {\bf W}^{\mathrm{yf}}y_{t-1} + b^{\mathrm{f}})\\
	o_t &=& \tilde{\sigma}({\bf W}^{\mathrm{xo}}x_t + {\bf W}^{\mathrm{yo}}y_{t-1} + b^{\mathrm{o}})\label{fourtheq}\\
	c_t &=& c_{t-1} \otimes f_t + i_t \otimes j_t\\
	y_t &=& \tanh(c_t) \otimes o_t \label{lasteq}
\end{eqnarray}
In these equations, $\bf W^*$ and $b^*$ correspond to learnable parameters (weights and biases, respectively) of the LSTM layer, and $\otimes$ corresponds to element-wise vector multiplication. $\tanh$ is the hyperbolic tangent function, and $\tilde{\sigma}$ is the \emph{hard sigmoid function}. To aid clarity, for the remainder of the model description, we will compress Equations \ref{firsteq}--\ref{lasteq} into $\mathrm{LSTM}(\vec{x}) = \vec{y}$,
representing a single LSTM layer, with its internal parameters and memory cell state kept implicit.

Our primary architecture represents a three-layer deep LSTM model for processing the historical weight/sleep/steps data. After performing the LSTM operations, the features of the final computed LSTM output step are concatenated with the height, gender, age category and weight objective, providing the following feature representation:
\begin{equation}
	\mathrm{LSTM}(\mathrm{LSTM}(\mathrm{LSTM}(\vec{\mathrm{wt}} || \vec{\mathrm{sl}} || \vec{\mathrm{st}})))_T || \mathrm{ht} || \mathrm{gdr} || \mathrm{age} || \mathrm{obj}
\end{equation}
where $\vec{\mathrm{wt}}$, $\vec{\mathrm{sl}}$ and $\vec{\mathrm{st}}$ are the input features (for weight, sleep and steps, respectively), $||$ corresponds to featurewise concatenation, and $T$ is the length of the initial sequence.
These features are passed through two fully connected neural network layers, connected to a single output neuron which utilises {a logistic sigmoid activation}.

The fully connected layers of the networks apply \emph{rectified linear} (ReLU) activations. We initialise the LSTM weights using Xavier initialisation \cite{glorot2010understanding}, and its forget gate biases with ones \cite{jozefowicz2015empirical}. Finally, the fully connected weights are initialised using He initialisation \cite{he2015delving}, as recommended for ReLUs. The models are trained for {200} epochs using the Adam SGD optimiser, with hyperparameters as described in \cite{kingma2014adam}, and a batch size of {1024}. For regularisation purposes, we have applied batch normalisation to the output of every hidden layer and {dropout with $p=0.1$ to the input-to-hidden transitions within the LSTMs} \cite{zaremba2014recurrent}.

\subsection{Cross-modal LSTM}

For this task we also propose a novel \emph{cross-modal} LSTM (\emph{X-LSTM}) architecture which {exploits} the \emph{multimodality} of the input data {more explicitly} in order to efficiently redistribute the LSTM's parameters. We initially partition the input sequence into three parts (sleep data, weight data, steps data), and pass \emph{each of those} through a separate three-layer LSTM stream. We also allow for \emph{information flow} between the streams in the second layer, by way of \emph{cross-connections}, where features from a single sequence stream are passed and concatenated with features from another sequence stream (after being passed through an additional LSTM layer). Represented via equations, the computed outputs across the three streams are: 
\begin{align}\label{e1}
	\vec{h}^\mathrm{wt}_1 &= \mathrm{LSTM}(\vec{\mathrm{wt}}) & \vec{h}^\mathrm{sl}_1 &= \mathrm{LSTM}(\vec{\mathrm{sl}}) & \vec{h}^\mathrm{st}_1 &= \mathrm{LSTM}(\vec{\mathrm{st}})\\
	\vec{h}^\mathrm{wt\rightarrow wt}_2 &= \mathrm{LSTM}(\vec{h}^\mathrm{wt}_1) & \vec{h}^\mathrm{sl\rightarrow sl}_2 &= \mathrm{LSTM}(\vec{h}^\mathrm{sl}_1) & \vec{h}^\mathrm{st\rightarrow st}_2 &= \mathrm{LSTM}(\vec{h}^\mathrm{st}_1)\\
	\vec{h}^\mathrm{wt\rightsquigarrow sl}_2 &= \mathrm{LSTM}(\vec{h}^\mathrm{wt}_1) & \vec{h}^\mathrm{sl\rightsquigarrow st}_2 &= \mathrm{LSTM}(\vec{h}^\mathrm{sl}_1) & \vec{h}^\mathrm{st\rightsquigarrow wt}_2 &= \mathrm{LSTM}(\vec{h}^\mathrm{st}_1)\\
	\vec{h}^\mathrm{wt\rightsquigarrow st}_2 &= \mathrm{LSTM}(\vec{h}^\mathrm{wt}_1) & \vec{h}^\mathrm{sl\rightsquigarrow wt}_2 &= \mathrm{LSTM}(\vec{h}^\mathrm{sl}_1) & \vec{h}^\mathrm{st\rightsquigarrow st}_2 &= \mathrm{LSTM}(\vec{h}^\mathrm{st}_1)
\end{align}
\begin{eqnarray}
	\vec{h}^\mathrm{wt}_3 &=& \mathrm{LSTM}(\vec{h}^\mathrm{wt\rightarrow wt}_2 || \vec{h}^\mathrm{sl\rightsquigarrow wt}_2 || \vec{h}^\mathrm{st\rightsquigarrow wt}_2)\\	
	\vec{h}^\mathrm{sl}_3 &=& \mathrm{LSTM}(\vec{h}^\mathrm{sl\rightarrow sl}_2 || \vec{h}^\mathrm{wt\rightsquigarrow sl}_2 || \vec{h}^\mathrm{st\rightsquigarrow sl}_2)\\	
	\vec{h}^\mathrm{st}_3 &=& \mathrm{LSTM}(\vec{h}^\mathrm{st\rightarrow st}_2 || \vec{h}^\mathrm{wt\rightsquigarrow st}_2 || \vec{h}^\mathrm{sl\rightsquigarrow st}_2)	\label{e2}
\end{eqnarray}

Finally, the feature representation passed to the fully connected layers is obtained by concatenating the final LSTM frames across all of the three streams: $(\vec{h}^\mathrm{wt}_3 || \vec{h}^\mathrm{sl}_3 || \vec{h}^\mathrm{st}_3)_T || \mathrm{ht} || \mathrm{gdr} || \mathrm{age} || \mathrm{obj}$

The illustration of the entire construction process from individual building blocks is {shown in} Figure \ref{figxlstm2}. This construction is biologically inspired by \emph{cross-modal systems} \cite{eckert2008cross} within the visual and auditory systems of the human brain---wherein several cross-connections between various sensory networks have been discovered \cite{Beer2011,yang2015}.

To provide breadth, we evaluate \emph{three} cross-connecting strategies: one as described by Equations \ref{e1}--\ref{e2} (\emph{A}), one where the cross-connection does not have intra-layer LSTMs (\emph{B}), and one where we don't cross-connect at all (\emph{N}). The latter corresponds the most to prior work on multimodal deep learning \cite{ngiam2011multimodal,srivastava2012multimodal} . Note that the variant (N) allows for computing the largest number of features within the parameter budget out of all three variants---no parameters being spent on cross-connections. The three scenarios are illustrated by Figure \ref{figconns}.

Finally, a recent state-of-the-art approach in processing multimodal sequential data \cite{Ren2016} imposes cross-modality by weight sharing between the different modalities' recurrent weights (${\bf W}^{y*}$ in Equations \ref{firsteq}--\ref{fourtheq})---we will refer to this technique as SH-LSTM. This comes at a cost to expressivity---in order to share them, these weight matrices need to have the same sizes, implying the different modality streams need to all compute the \emph{same number of features} at each depth level. Keeping the parameter count comparable to the baseline LSTM, we evaluate three strategies for weight sharing (Figure \ref{figconns}): sharing across all modalities (ALL) and sharing across weight/sleep only, with (WSL) and without (CUT) steps data. This has been motivated by the fact that the weight and sleep data have, on their own, been found to be significantly more influential than steps data---as will be discussed in the Results section.

\begin{figure}
	\centering
	\begin{sideways}
  \includegraphics[width=0.5\linewidth]{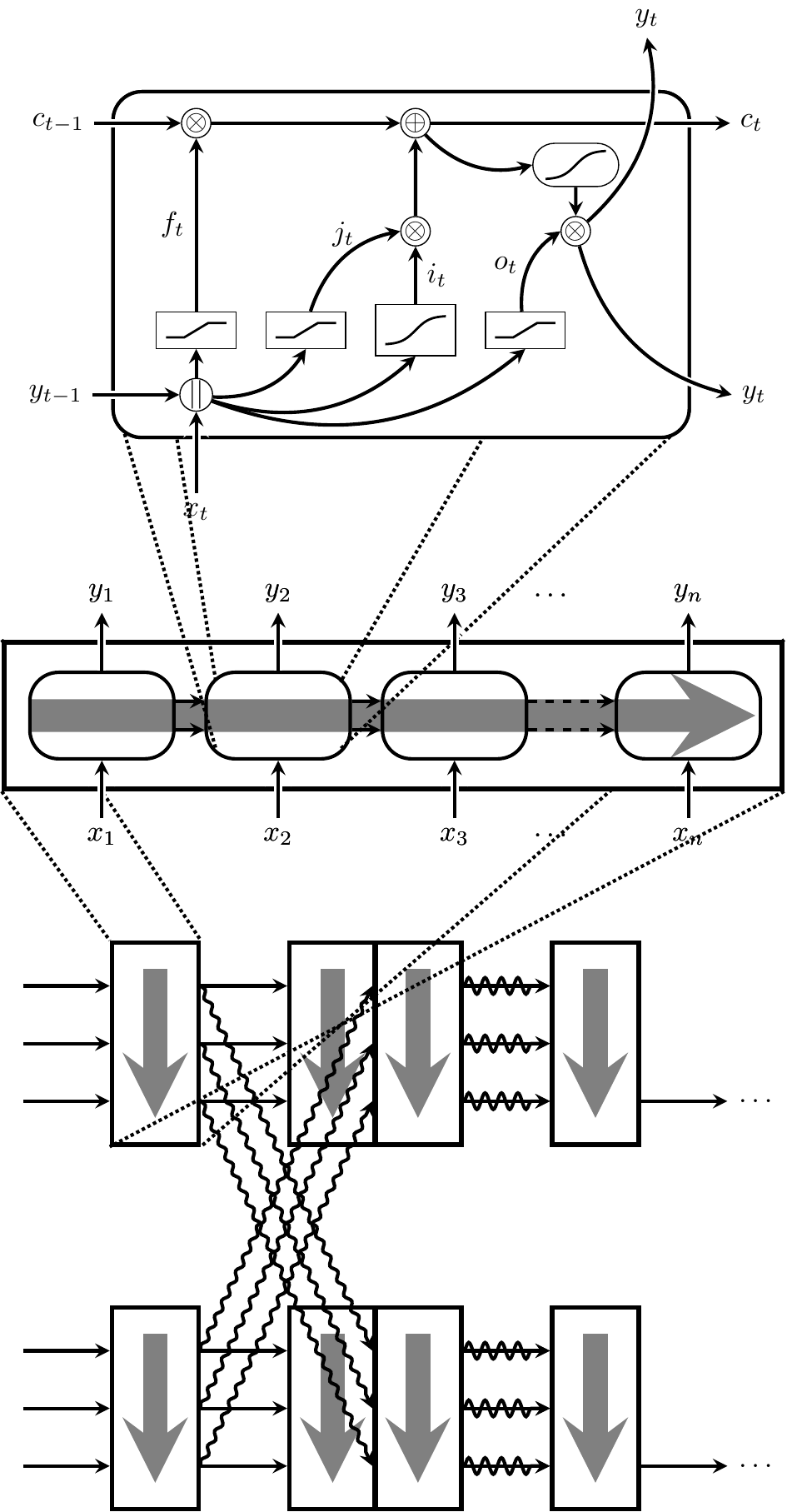}
	\end{sideways}
	\caption{A hierarchical illustration of a deep X-LSTM model with three layers and one cross-connection in the second layer. \textbf{Left:} A single LSTM block; all intermediate results, as described in Equations \ref{firsteq}--\ref{lasteq} ($i_t$, $j_t$, $f_t$ and $o_t$) are clearly marked. \textbf{Middle:} Replicating the LSTM cell to create an LSTM layer (for processing a given input sequence $\vec{x}$). \textbf{Right:} A cross-modal deep LSTM model with two streams of three layers, taking sequences of length 3. In the second layer, the hidden sequences are shared between the two streams by being passed through a separate LSTM layer and feature-wise concatenated with the main stream hidden sequence.}
    \label{figxlstm2}
\end{figure}

\begin{figure}
\centering
\includegraphics[height=0.25\linewidth]{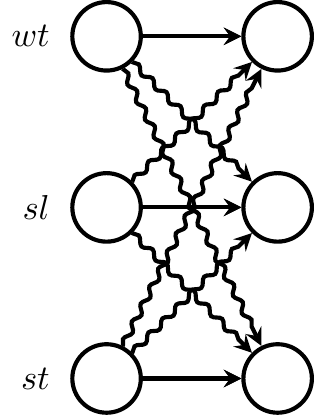}\hfill
\includegraphics[height=0.25\linewidth]{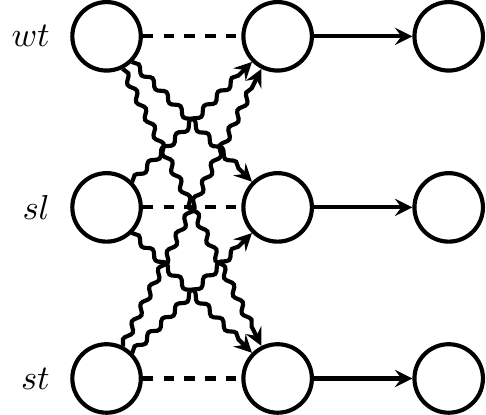}\hfill
\includegraphics[height=0.25\linewidth]{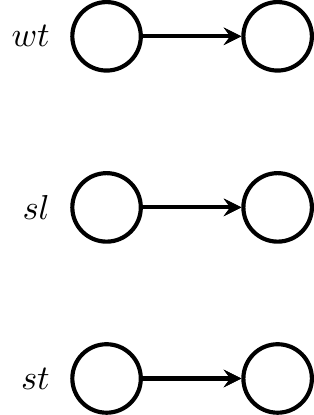}\\[4ex]
\includegraphics[height=0.25\linewidth]{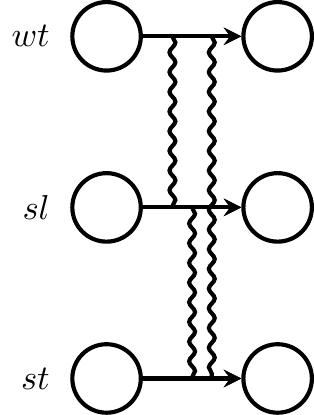}\hfill
\includegraphics[height=0.25\linewidth]{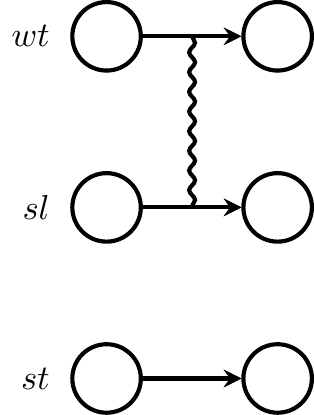}\hfill
\includegraphics[height=0.25\linewidth]{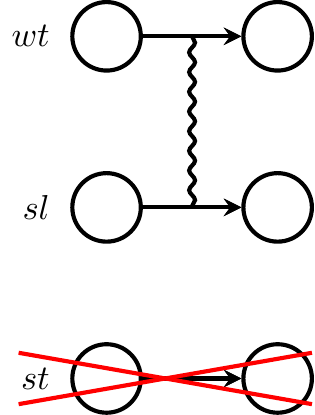}
\caption{The three types of cross-connection strategies, and the three types of weight sharing strategies. Each arrow is an LSTM layer, dashed lines indicate the identity transformation, and all arrows going into the same node are concatenated. Connections between lines in the bottom row represent recurrent weight sharing. {\bf Top, left-to-right:} X-LSTM (A), X-LSTM (B), X-LSTM (N). {\bf Bottom, left-to-right:} SH-LSTM (ALL), SH-LSTM (WSL), SH-LSTM (CUT).}
\label{figconns}
\end{figure}

\subsection{Weight objective success classification}

We performed stratified 10-fold crossvalidation on the baseline classifiers as well as the proposed LSTM model. Given the bias of the obtained data towards failure (there being \emph{twice as many} sequences labelled unsuccessful), and the fact that it is not generally obvious what the classification threshold for this task {should be} (it likely involves several tradeoffs), we use \textbf{ROC curves} (and the associated \emph{area} under {them}) as our primary evaluation metric. For completeness, we also report the accuracy, precision, recall, F$_1$ score and the Matthews Correlation Coefficient \cite{matthews1975comparison} under the classification threshold {which maximises the F$_1$ score}. 

Afterwards we sought to construct competitive X-LSTMs, and therefore we computed the AUCs of the individual unimodal LSTMs on a validation dataset, obtaining AUCs of $80.62\%$ (for weight), $80.17\%$ (for sleep) and $74.18\%$ (for steps). As anticipated, this was not far enough in order to reliably generate non-uniform X-LSTMs, so we proceeded to perform a grid search on the parameter $k$. We've originally taken steps of $5$, but as we found the differences between adjacent steps to be negligible, we report the AUC results for $k \in \{10, 20, 30\}$. The X-LSTM performed the best with $k = 30$, and (B) cross-connections---we compare it directly with the LSTM, as well as the SH-LSTMs, and report its architecture in Table \ref{tblarch}.

\begin{table}
    \centering
    \small{
	\begin{tabular}{c c} 
    \toprule
    	LSTM & X-LSTM (B, $k = 30$)\\ 
        {76377} param. & {75089} param.\\ \midrule
        {21} features & wt: {15} features, sl: {12} features, st: {2} features\\
        &{\bf wt $\rightsquigarrow $ sl: {9} features, wt $\rightsquigarrow $ st: {14} features}\\
        & {\bf sl $\rightsquigarrow $ wt: {6} features, sl $\rightsquigarrow $ st: {11} features}\\
        & {\bf st $\rightsquigarrow $ wt: {1} feature, st $\rightsquigarrow $ sl: {1} feature}\\
        {42} features & wt: {29} features, sl: {24} features, st: {3} features\\
        {84} features & wt: {57} features, sl: {48} features, st: {5} features\\ \midrule
        \multicolumn{2}{c}{Fully connected, {128}-D} \\
        \multicolumn{2}{c}{Fully connected, {64}-D} \\ 
        \multicolumn{2}{c}{Fully connected, 1-D}\\
        \bottomrule
    \end{tabular}
    }
    \caption{Architectures for the considered LSTM and cross-modal LSTM models. Cross-connections are {\bf highlighted}.}
\label{tblarch}
\end{table}

{To confirm that the advantages demonstrated by our methodology are statistically significant, we have performed paired $t$-testing on the metrics of individual cross-validation folds, choosing a significance threshold of $p < 0.05$. We find that all of the observed advantages in ROC-AUC are indeed statistically significant---verifying simultaneously that the recurrent models are superior to other baseline approaches, that the X-LSTM has significantly improved on its LSTM baselines and that cross-connecting is statistically beneficial (given the weaker performance of X-LSTM (N) despite being able to compute the most features overall).} The SH-LSTM performed the best in its (WSL) variant (which allowed for more features to be allocated to weight and sleep streams, at the expense of the steps stream) but was even then unable to outperform the baseline LSTM---highlighting once again its lack of ability to accurately specify relative importances between modalities, which is essential for this task. The results are summarised by Tables \ref{tblbaseline}--\ref{tblcls} and Figure \ref{figrocsbase}.

\begin{table*}
\centering
\small{
\begin{tabular}{l c c c c c c c} \toprule
{\bf Metric} & SVM & RF & GHMM & DNN & LSTM & SH-LSTM & X-LSTM\\ \midrule
Accuracy & 67.65\% &	70.97\% &	66.31\% &	68.93\% & 79.12\% & 78.49\% & {\bf 80.30\%} \\
Precision & 52.54\% &	56.05\% &	51.26\% &	53.80\%	& 67.25\% & 65.31\% & {\bf 68.66\%} \\
Recall & 81.02\% &	81.34\% &	82.32\% &	{\bf 83.02\%} &	79.30\% & 82.95\% & 81.62\% \\
F$_1$ score & 63.71\% &	66.25\% &	63.11\% &	65.18\% &	72.69\% & 72.98\% & {\bf 74.37\%} \\
MCC & 39.74\% &	44.75\% &	38.57\% &	42.63\% &	 56.60\% & 56.80\% & {\bf 59.45\%} \\ \midrule
\underline{ROC AUC}& 76.77\% & 79.97\% & 74.86\% &	78.54\%	& 86.91\% & 86.63\% & {\bf 88.07\%}\\ 
$p$-value & \underline{$2 \cdot 10^{-12}$} & \underline{$6 \cdot 10^{-10}$} & \underline{$7 \cdot 10^{-11}$} & \underline{$2 \cdot 10^{-11}$} & \underline{$1 \cdot 10^{-4}$} & \underline{$4 \cdot 10^{-5}$} & ---\\\bottomrule
\end{tabular}
}
\caption{Comparative evaluation results of the baseline models against the LSTMs after 10-fold crossvalidation. Reported X-LSTM is the best-performing (B, $k=30$). Reported SH-LSTM is the best-performing (WSL). All metrics except the ROC AUC reported for the classification threshold that maximises the F$_1$ score. Reported $p$-values are for the X-LSTM vs. each baseline for the ROC-AUC metric.}
\label{tblbaseline}
\end{table*}

\begin{figure}
\centering
\includegraphics[width=0.6\linewidth]{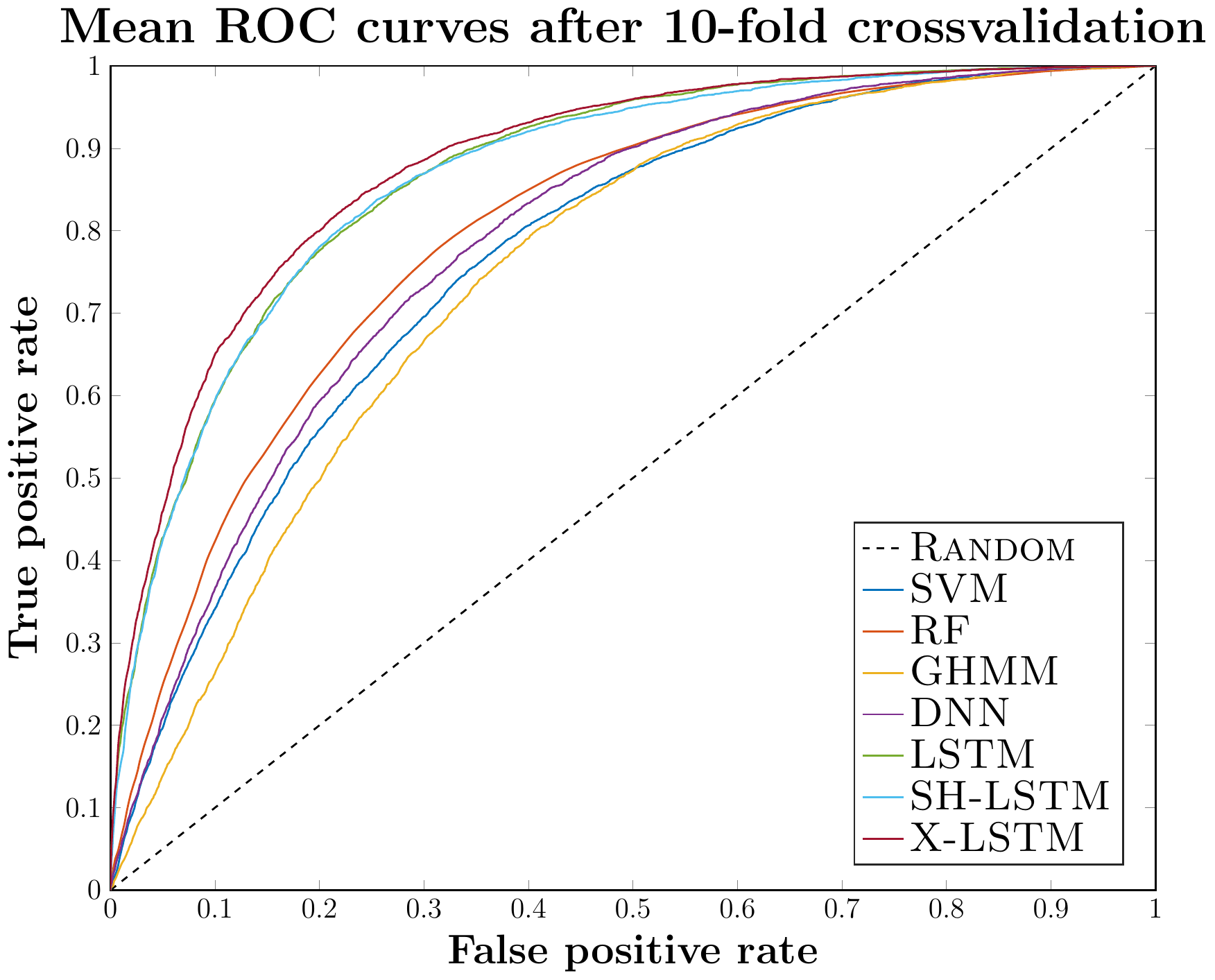}
\caption{Mean ROC curves for the baselines, LSTM and the best-performing SH-LSTM and X-LSTM models.}
\label{figrocsbase}
\end{figure}

\begin{table}
\centering
\small{
\begin{tabular}{l c c c} \toprule
{\bf Model} & $k=10$ & $k=20$ & $k=30$\\ \midrule
X-LSTM (A) & {\bf 87.60\%} &	 {\bf 87.60\%} &	87.75\%\\
X-LSTM (B) & 87.21\% &	87.56\% &	{\bf 88.07\%}\\
X-LSTM (N) & 86.49\% &	86.98\% &	87.30\%\\
$p$-value & \underline{$9.55 \cdot 10^{-5}$} & \underline{$0.021$}& \underline{$1.03 \cdot 10^{-3}$}\\\midrule
SH-LSTM (ALL) & \multicolumn{3}{c}{85.58\%}\\
SH-LSTM (WSL) & \multicolumn{3}{c}{86.63\%}\\
SH-LSTM (CUT) & \multicolumn{3}{c}{86.30\%}\\\bottomrule
\end{tabular}
}
\caption{Effects of varying the hyperparameter $k$ and cross-connecting strategy of X-LSTMs to the mean ROC AUC after crossvalidation. Reported $p$-values are for the (N) vs. $\max$(A, B) strategies. We also report the mean ROC AUC for the three kinds of sharing strategies of SH-LSTMs.}
\label{tblcls}
\end{table}

\subsection{{Weight objective magnitude effects}}

\begin{figure}
\centering
\includegraphics[width=0.49\linewidth]{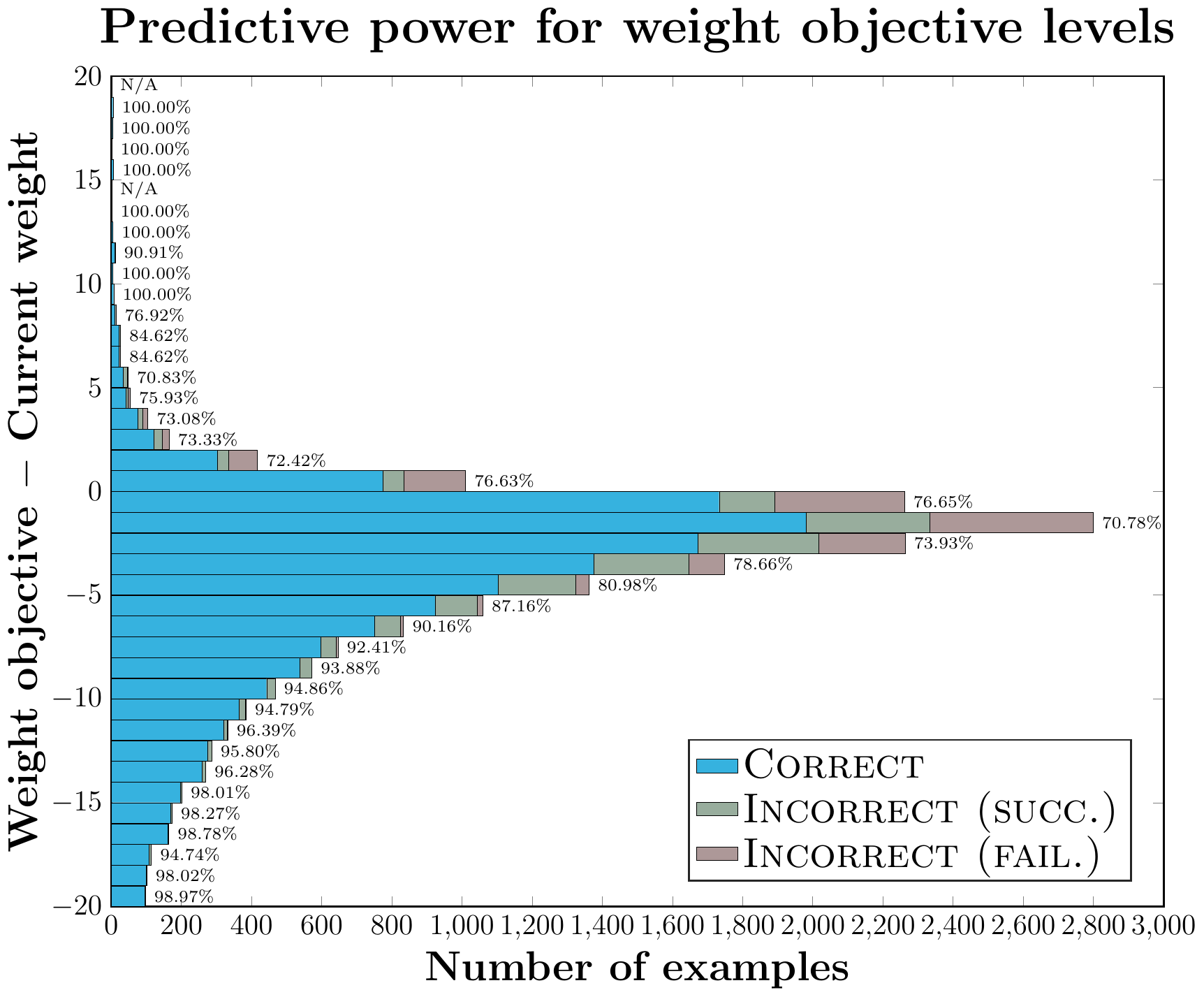}
\includegraphics[width=0.49\linewidth]{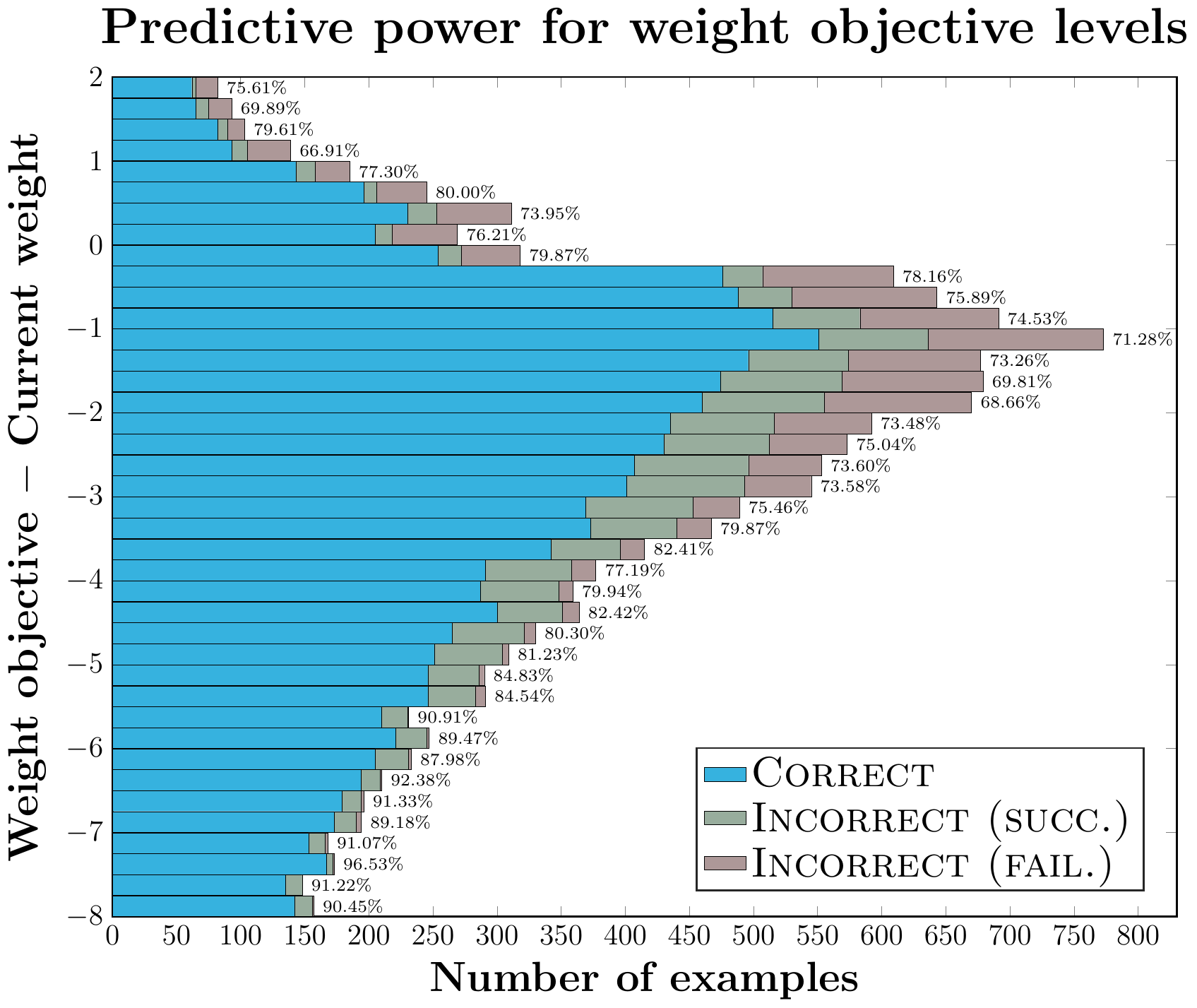}
\caption{{{\bf Left:} A bar plot demonstrating the X-LSTM's performance for different magnitudes of weight objectives (at the classification threshold of 0.5). {\bf Right:} The same plot, zoomed in on the $[-8, 2]$ range of weight objectives (where the majority of the examples are).}}
\label{figbars}
\end{figure}

{The \emph{magnitude} of weight objectives set by users will have an obvious impact on the predictive power of the model. To illustrate this effect on the X-LSTM, we have aggregated its predictions across all of the crossvalidation folds (for a classification threshold of 0.5) into a histogram using bins of various weight objective magnitude ranges (ref. Figure }\ref{figbars}{). The histogram shows the proportion of correctly classified, incorrectly classified successful and incorrectly classified failed sequences.}



{The results closely match our expectations---at smaller weight objective magnitudes, the model is unbiased towards success or failure. However, starting at $-3  \mathrm{kg}$ and moving higher, there is a clear bias towards misclassifying successful sequences, which eventually grows into nearly all misclassified sequences being successful. This kind of behaviour is fairly desirable---as it will encourage selection of realistic objectives, at the expense of making incorrect initial predictions about a few users that do eventually manage to achieve very ambitious goals.}

\end{document}